\documentclass{article} 
\usepackage{colm2024_conference}

\usepackage{microtype}
\usepackage{hyperref}
\usepackage{url}
\usepackage{booktabs}
\usepackage{enumitem}
\usepackage{latexsym}
\usepackage{graphicx}
\usepackage{wrapfig}
\usepackage{multirow} 
\usepackage{multicol} 
\usepackage{makecell}
\usepackage{hhline}
\usepackage{pifont}
\usepackage{cleveref}
\usepackage{amssymb}
\usepackage{subcaption}
\usepackage{tabularx}

\title{MARVEL: Multidimensional Abstraction and Reasoning through Visual Evaluation and Learning}

\author{Yifan Jiang$^1$\thanks{Authors contributed equally}~~\quad
    Jiarui Zhang$^{1*}$\quad
    Kexuan Sun$^{1*}$ \quad
    {\bf Zhivar Sourati$^1$}\quad \\
    {\bf Kian Ahrabian$^1$}\quad
    {\bf Kaixin Ma$^2$}\quad
    {\bf Filip Ilievski$^3$}\quad    
    {\bf Jay Pujara$^1$}\\
$^1$Information Sciences Institute, University of Southern California \\
$^2$Tencent AI Lab, Bellevue, WA\\
$^3$Department of Computer Science, Faculty of Science, Vrije Universiteit Amsterdam\\
    {\small \texttt{\{yjiang44,jzhang37,kexuansu,souratih,ahrabian\}@usc.edu}}\\
    {\small \texttt{kaixinma@global.tencent.com} , \quad \texttt{f.ilievski@vu.nl}, \quad \texttt{jpujara@isi.edu}}
}

\colmfinalcopy 
\begin{document}
\definecolor{darkergreen}{RGB}{50,160,50} 
\definecolor{lightblue}{HTML}{70f3ff}
\definecolor{lightgreen}{HTML}{00e09e}
\definecolor{checkmark}{HTML}{2edfa3}
\newcommand{\jiarui}[1]{{\color{darkergreen}(JR: #1)}} 
\newcommand{\kian}[1]{{\color{red}(KA: #1)}}   
\newcommand{\yifan}[1]{{\color{purple}(YF: #1)}} 
\newcommand{\kexuan}[1]{{\color{orange}(KX: #1)}} 
\newcommand{\zhivar}[1]{{\color{blue}(ZS: #1)}} 
\newcommand{\filip}[1]{{\color{lightblue}(FI: #1)}} 
\newcommand{\kaixin}[1]{{\color{lightgreen}(KM: #1)}} 
\newcommand{\cm}{\textcolor{checkmark}{\ding{52}}}

\newcommand{\ncm}{\textcolor{yellow}{\ding{52}}}
\newcommand\blfootnote[1]{%
  \begingroup
  \renewcommand\thefootnote{}\footnote{#1}%
  \addtocounter{footnote}{-1}%
  \endgroup
}

\maketitle
\begin{abstract}
While multi-modal large language models~(MLLMs) have shown significant progress on many popular visual reasoning benchmarks, whether they possess abstract visual reasoning abilities remains an open question. Similar to the Sudoku puzzles, abstract visual reasoning~(AVR) problems require finding high-level patterns~(e.g., repetition constraints) that control the input shapes~(e.g., digits) in a specific task configuration~(e.g., matrix). However, existing AVR benchmarks only considered a limited set of patterns~(addition, conjunction), input shapes~(rectangle, square), and task configurations~($3 \times 3$ matrices). To evaluate MLLMs' reasoning abilities comprehensively, we introduce \textbf{\textsc{MARVEL}}, a multidimensional AVR benchmark with 770 puzzles 
composed of six core knowledge patterns, geometric and abstract shapes, and five different task configurations. To inspect whether the model accuracy is grounded in perception and reasoning, \textsc{MARVEL} complements the general AVR question with \textit{perception questions} in a hierarchical evaluation framework. We conduct comprehensive experiments on \textsc{MARVEL} with nine representative MLLMs in zero-shot and few-shot settings. Our experiments reveal that all models show near-random performance on the AVR question, with significant performance gaps~(40\%) compared to humans across all patterns and task configurations. Further analysis of perception questions reveals that MLLMs struggle to comprehend the visual features~(near-random performance) and even count the panels in the puzzle~(\textless45\%), hindering their ability for abstract reasoning. We release our entire code and dataset.\footnote{\url{https://github.com/1171-jpg/MARVEL\_AVR}} 

\end{abstract}

\section{Introduction}

\begin{figure}[t]
  \centering
  \includegraphics[width=0.9\textwidth]{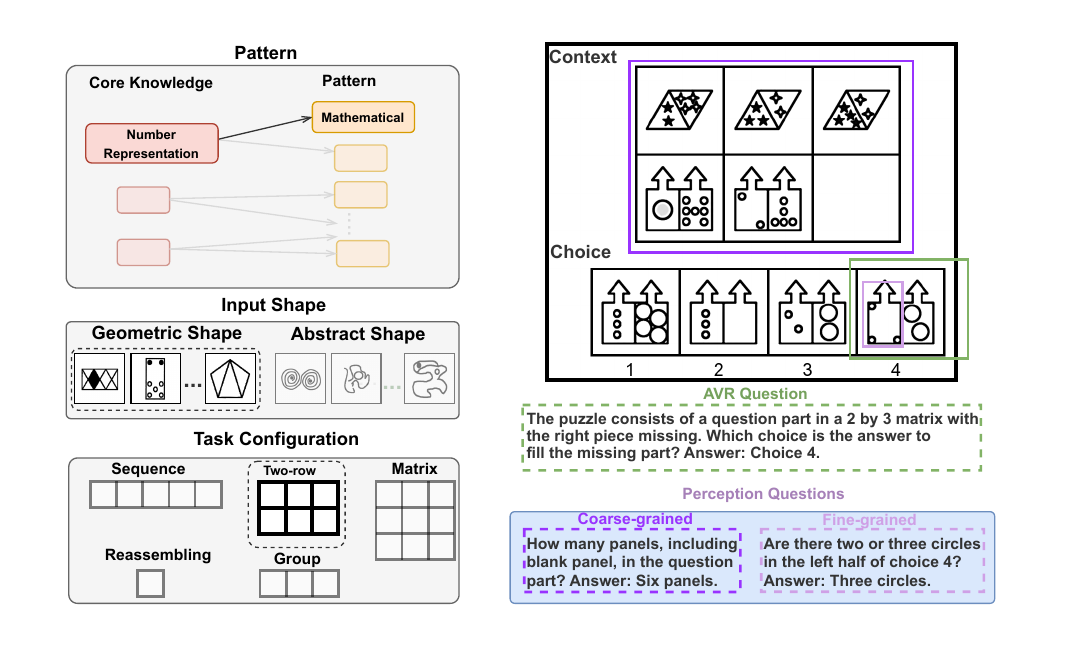}
  \caption{An abstract visual reasoning puzzle in \textbf{MARVEL}. The puzzle contains \textbf{mathematical} pattern governing the element number in \textbf{geometric shapes} with \textbf{two-row} task configuration. The \textcolor{teal}{AVR question}  focuses on the final answer for the puzzle, while the \textcolor{violet}{perception questions} focus on the fine-grained detail about one choice or coarse-grained detail over the whole puzzle.}
  \label{fig:example}
  \vspace{-0.3cm}
\end{figure}

Recent advances in novel training pipelines, computational resources, and data sources have enabled Multi-modal Large Language Models~(MLLMs)~\citep{openai2023gpt,geminiteam2023gemini} to show strong visual reasoning ability in tasks that require both visual and textual cues~\citep{wang2023review}, such as visual question answering~\citep{goyal2017making,antol2015vqa} and visual commonsense reasoning~\citep{zellers2019recognition,xie2019visual}. These tasks typically focus on testing the models' real-world knowledge~\citep{malkinski2023review}.
On the other hand, abstract visual reasoning~(AVR)~\citep{zhang2019raven,hill2019learning} has little dependency on the world knowledge. As the puzzle shown in \autoref{fig:example} (top-right), AVR problems require identifying the hidden pattern~(addition and subtraction) that governs the input shapes and their attribute~(number of stars/circles) in a task configuration~(2 $\times$ 3 matrix). 
The abstract reasoning ability is related to many practical applications, including visual representations~\citep{patacchiola2020self} and anomaly detection~\citep{schubert2014local}, encouraging more fundamental research in evaluating MLLMs on AVR benchmarks~\citep{ahrabian2024curious,yang2023dawn}.

However, the scope of current AVR benchmarks is limited, covering few reasoning patterns over a limited set of input shapes arranged in a predetermined configuration of puzzle panels, leading to biased evaluations~\citep{van2021much}. 
RAVEN~\citep{zhang2019raven} tests abstract reasoning mostly in mathematical patterns over predefined geometric input shapes. Bongard-LOGO~\citep{nie2020bongard} and SVRT~\citep{fleuret2011comparing} implement geometric-related~(symmetric) patterns over manually designed abstract input shapes.
Although recent surveys highlight the limited evaluation scope of these benchmarks~\citep{malkinski2023review,van2021much}, a comprehensive benchmark for accessing MLLMs in multidimensional settings has not been proposed and is still needed.

To fill this gap, we introduce \textbf{\textsc{MARVEL}}, a multi-dimensional abstract visual reasoning benchmark designed to evaluate MLLMs across different patterns, input shapes, and task configurations.
\textsc{MARVEL}'s underlying reasoning patterns are rooted in key core knowledge of human cognition, observed in newborn infants relying only on abstraction to reason about their environment~\citep{spelke2007core}. 
To facilitate a more comprehensive evaluation of foundation MLLMs, \textsc{MARVEL} combines six patterns expanded from three types of core knowledge, both geometric and abstract shapes, and five task configurations.
We crawl relevant puzzles from publicly available websites, manually filter low-quality and irrelevant puzzles based on the expanded patterns and input shapes, and reformat them into different task configurations. 
We annotate the AVR question by providing a short description of the puzzle and asking for its answer. In total, we collect 770 diverse and high-quality puzzles assessing abstract visual reasoning abilities~(\autoref{fig:example}). 

To provide a hierarchical evaluation, we also enrich each puzzle with perception questions focusing on perceiving puzzles’ visual details (e.g., number of grids, edges of a triangle) to measure
models’ reasoning consistency~\citep{jiang2023brainteaser,selvaraju2020squinting}.
We conduct comprehensive experiments on MARVEL involving different model structures, model sizes, and prompting strategies. Our experiments reveal that all MLLMs show near-random performance in all patterns, even with few-shot demonstrations and prompt engineering, leaving a huge gap~(40\%) in the abstract reasoning ability with humans. An in-depth analysis based on perception questions points out that MLLMs' poor reasoning ability is hindered by their struggle with fine-grained visual feature comprehension. Their near-random perception ability fails to provide foundations for subsequent abstract reasoning.
In summary, our contributions can be listed as follows: 1) \textbf{A novel multidimensional AVR benchmark}, \textbf{\textsc{MARVEL}}, which consists of six patterns across five distinct task configurations. 2) \textbf{A hierarchical evaluation framework} incorporating perception questions with AVR questions to enable fine-grained diagnosis of model capability. 3) \textbf{Extensive experiments} on a wide range of state-of-the-art MLLMs with various prompting strategies, providing insights into their strengths, weaknesses, and future improvement directions on AVR tasks and MLLM developments.

\begin{table}[t]
\centering
\resizebox{\textwidth}{!}{%
\begin{tabular}{ll|c|c|c|c|c|c|c|c}
\toprule
\multicolumn{2}{c|}{Dimension}                                            & RAVEN & G-set& VAP & \makecell[c]{Bongard-\\ LOGO}& SVRT &  DOPT & ARC* & \textbf{MARVEL} \\ \midrule
\multicolumn{1}{l}{\multirow{2}{*}{\textbf{Input Shape}}}   & Geometric            & \cm  & \cm   & \cm &                              &      &\cm  & \cm & \cm    \\
\multicolumn{1}{l}{}                               & Abstract             &      &       &     &      \cm                              & \cm  &     &     & \cm     \\ \midrule
\multicolumn{1}{l}{\multirow{6}{*}{\textbf{Pattern}}}       & Temporal Movement    & \cm  & \cm   &     &                              &      & \cm & \cm & \cm     \\
\multicolumn{1}{l}{}                               & Spatial Relationship &      &       &     &                                       & \cm   &     & \cm  & \cm     \\
\multicolumn{1}{l}{}                               & Quantities          & \cm  & \cm   & \cm &     \cm                           & \cm   & \cm  & \cm  & \cm     \\
\multicolumn{1}{l}{}                               & Mathematical         & \cm  & \cm    & \cm &                                     &      &     & \cm & \cm     \\
\multicolumn{1}{l}{}                               & 2D-Geometry          &      &       &     &      \cm                              & \cm  &     &     & \cm    \\
\multicolumn{1}{l}{}                               & 3D-Geometry          &      &       &     &                                       &      &     &     & \cm   \\ \midrule
\multicolumn{1}{l}{\multirow{5}{*}{\textbf{Configuration}}} & Sequence             &      &       &     &                              &      & \cm &     & \cm   \\
\multicolumn{1}{l}{}                               & Two-row              &      &       & \cm &                                       &      &     &     & \cm    \\
\multicolumn{1}{l}{}                               & Matrix               & \cm  & \cm   &     &                                       &      &     &     & \cm   \\
\multicolumn{1}{l}{}                               & Group                &      &       &     &     \cm                              &  \cm  &     &     & \cm    \\
\multicolumn{1}{l}{}                               & Reassembling           &      &       &     &                                        &      &     &     &  \cm   \\ \midrule
\multicolumn{2}{c|}{Perception Question}                  &      &       &     &                              &      &     &     &  \cm \\ \bottomrule
\end{tabular}%
}
\caption{Comparing \textsc{MARVEL} to related benchmarks: RAVEN~\citep{zhang2019raven}, G-set~\citep{mandziuk2019deepiq}, VAP~\citep{hill2019learning}, Bongard-LOGO~\citep{nie2020bongard}, SVRT~\citep{fleuret2011comparing}, ARC~\citep{chollet2019measure}, DOPT~\citep{webb2020learning}. *ARC puzzles are provided in a generative format.}
\label{tab:dataset}
\vspace{-0.3cm}
\end{table}

\section{Related Work}

\paragraph{MLLM Evaluations.}
Benefiting from the rich representation from visual encoders~\citep{clip} and strong reasoning ability of LLMs~\citep{llama2,vicuna},
MLLMs~\citep{li2023blip,dai2024instructblip,openai2023gpt4,llava} have been applied to solve not only traditional vision-language tasks, such as image captioning~\citep{nocaps,flick}, visual question answering~\citep{vqav2,okvqa,gqa,textvqa} and refer expression comprehension~\citep{refcoco,grit}, but also on more complicated scenarios, such visually-grounded conversation~\citep{llava,flamingo}, multimodal web/UI agents~\citep{he2024webvoyager,zheng2024gpt4vision,appagent} and embodied tasks~\cite{palme}. Besides end-to-end evaluation, several recent works also try to reveal MLLMs' visual shortcomings from different aspects, including visual details~\citep{vicrop,vstar}, perceptual bias~\citep{perceptuallimitation}, and small visual pattern recognition~\citep{eyeswideshut}. Although some of the existing benchmarks have accessed MLLM's mathematical visual reasoning abilities requiring an understanding of abstract and geometry shapes~\citep{lu2023mathvista,scienceqa}, their evaluation still heavily relies on textual descriptions. In contrast, AVR benchmarks assess MLLMs' ability under a diverse set of patterns with only visual understanding settings.

\paragraph{AVR Benchmarks.}
AVR problems have great potential impact on various domains~\citep{malkinski2023review,patacchiola2020self,schubert2014local}, sparking interests in evaluating MLLMs on AVR benchmarks~\citep{ahrabian2024curious,moskvichev2023conceptarc,mitchell2023comparing}. 
Existing AVR benchmarks present the evaluation in a wide range of formats, such as selective completion~\citep{zhang2019raven,hu2021stratified,benny2021scale,webb2020learning}, group discrimination~\citep{fleuret2011comparing,nie2020bongard} and generative completion~\citep{chollet2019measure}. However, less attention is paid to the scope and pattern of the AVR benchmark; most focus only on a few simple abstract patterns and testing models end-to-end without considering the intermediate perception and reasoning procedures~\citep{moskvichev2023conceptarc,mitchell2021abstraction}. In contrast, MARVEL includes geometric and abstract shapes, six core patterns essential for visual abstract reasoning, and five different task configurations. Inspired by prior analysis of visual details and perceptual bias, MARVEL introduces perception questions to ensure the MLLMs correctly perceive the presented visual patterns. Our work and other related AVR benchmarks are compared in~\Cref{tab:dataset}.

\section{MARVEL Benchmark Construction}
As a multidimensional benchmark for AVR, \textsc{MARVEL} covers different task configurations~(\Cref{subsec:task-definition-and-configuration}), various input shapes~(\Cref{subsec:input-shapes}), as well as different reasoning patterns involved in the puzzles~(\Cref{subsec:rule}). Further, \textsc{MARVEL} decomposes the evaluation of models' capabilities into 1) perception questions~(\Cref{sec: evaluation_framework}) about the puzzle panels and 2) AVR questions~(\Cref{subsec:data-collection-analysis}) to yield a more precise and faithful picture of evaluated models. The data collection process is presented in \Cref{subsec:data-collection-analysis}.

\subsection{Task Definition and Configurations}
\label{subsec:task-definition-and-configuration}
Each puzzle in \textsc{MARVEL} consists of a context on the top and possible choices ($c_i; i \in \{1,2,3,4\}$) to choose from at the bottom, formatted in a multiple-choice question answering setting (see \Cref{fig:example} for an example).
The context part consists of $n$ puzzle panels ($p_1,p_2,\dots,p_n,p_b$ with $p_b$ being a blank panel), with their specific number and arrangement driven by a task configuration and reasoning pattern, $P$, that governs the relationship between puzzle panels. The choice will be considered the correct answer and fill in $p_b$ that can satisfy the following equation: 
$P(p_1,p_2,\dots,p_n) = P(p_1,p_2,\dots,p_n,c_c)$.

Puzzle panels in \textsc{MARVEL} are organized in the following five task configurations:

\begin{enumerate}[leftmargin=*,itemsep=1pt,topsep=0pt,parsep=0pt]
\item \textbf{Sequence Format} arranges panels in a $1 \times n$ line~($n \in [4,7]$).
\item \textbf{Two-row Format} presents panels in a $2 \times 3$ matrix$^*$\blfootnote{*$p_a^b$: the panel on the $a$ th row and $b$ th column of matrix.}, $p_1^1,p_2^1,p_3^1$ and $p_1^2,p_2^2,p_3^2$. 
The solution requires identifying the same pattern at the first row and the second row, which is $ P(p_1^1,p_2^1,p_3^1) = P(p_1^2,p_2^2,c_c)$.

\item \textbf{Matrix Format} organizes panels in a 3 by 3 matrix, the pattern can be reflected in either row- or column-wise way: $ P(p_1^1,p_2^1,p_3^1) =P(p_1^2,p_2^2,p_3^2) = P(p_1^3,p_2^3,c_c)$ or $ P(p_1^1,p_1^2,p_1^3) = P(p_2^1,p_2^2,p_3^2) = P(p_3^1,p_3^2,c_c)$.



\item \textbf{Group Format} has three panels in the question part~($p_1,p_2,p_b$) with one choice reflecting the context's pattern and other choices differing: $ P(p_1,p_2,c_c) \neq P^\prime(choices - c_c)$.

\item \textbf{Reassembling Format} is designed for \textit{3D-Geometry} pattern with one unfolded diagram in the context and four 3D assembly results as choices.
\end{enumerate}
We visualize these configurations in ~\Cref{fig:example} (see detailed examples in~\Cref{app: examples}).
\subsection{Input Shapes}
\label{subsec:input-shapes}

As shown in Figure~\ref{fig:example}, each panel of a puzzle contains various shapes that can be generally differentiated into two types~\citep{malkinski2023review}: 
\begin{enumerate}[leftmargin=*,itemsep=1pt,topsep=0pt,parsep=0pt]
 \item \textbf{Geometric Shapes} are easily comprehended and described. For instance, a square is a shape that has four sides of equal length and four equal angles. Most existing AVR benchmarks~\citep{zhang2019raven,hill2019learning} focus on elementary shapes such as oval, rectangle, triangle and trapezoid. \textsc{MARVEL} includes geometric shapes consisting of more than two different elementary geometric shapes to mitigate the issue and improve the complexity.
 \item \textbf{Abstract Shapes} come from a wide set of possibilities and vary widely from one problem to another. It provides a fair step as most MLLMs encounter the shapes for the first time. \textsc{MARVEL} also includes abstract shapes as they are gaining more preference for AVR-related research~\citep{fleuret2011comparing,nie2020bongard}.
 \end{enumerate}



\subsection{Core Knowledge and Patterns}
\label{subsec:rule}

Core knowledge theory~\citep{spelke2007core} from cognition developmental psychology is largely shared among humans and particularly for human infants. Human infants with no real-world knowledge and limited experience represent their environment using abstraction patterns. These abstraction patterns can be categorized into three types of core knowledge, which is the foundation for inference and reasoning~\citep{lipton2004discrimination}. We expand each core knowledge\footnote{We don't feature the fourth Core Knowledge, agent representation, due to its concentrates on goal-directed and interactive action, which is not adaptable in \textsc{MARVEL} setting.} into two patterns for a fine-grained assessment of abstract reasoning in \textsc{MARVEL}, based on insights drawn from contemporary cognitive literature:

\textbf{Object Core Knowledge} represents objects' spatio-temporal motions and their contact, which is expanded to \textit{Temporal Movement Pattern} focusing on the related position change or movement~\citep{spelke1990principles} and \textit{Spatial Relationship Pattern} examining objects' relative positional relationship~\citep{aguiar19992}.

\textbf{Number Core Knowledge} helps infants process abstract representations of small numbers. We include \textit{Quantities Pattern} testing the accuracy of number comprehension~\citep{xu2000large} and \textit{Mathmatical Pattern} for elementary mathematical operations~\citep{barth2005abstract}.

\textbf{Geometry Core Knowledge} captures the environment's geometry, which helps humans orient themselves in their surroundings. We divide it into \textit{2D-Geometry Pattern}~\citep{hermer1996modularity} and \textit{3D-Geometry Pattern}~\citep{cheng2005there}.

\subsection{Data Collection}
\label{subsec:data-collection-analysis}

We collect puzzles from several public resources websites\footnote{\url{https://www.gwy.com/}; \url{https://www.chinagwy.org/}} and filter out unfit or low-quality data by three human annotators based on the puzzle's input shapes~(some puzzles contain textual information) and patterns. Unaligned puzzles are first segmented into panels and then reassembled into the correct task configuration. To ensure each pattern in each task configuration has at least 45 puzzles\footnote{Some patterns can not be presented in specific configurations. For example, the \textit{Mathmatical Pattern} can not be adapted to group format as it requires comparison between adjacent panels.}, we also manually created 220 puzzles by following the pattern in existing data and replacing the input shape drawn from scratch. Each puzzle contains an AVR question~(\autoref{fig:example}) generated from templates based on their task configuration. AVR questions provide a brief description and ask only for the puzzle's final answer, which is widely adopted in previous AVR benchmark~\citep{malkinski2023review}.
In the end, \textsc{MARVEL} includes 770 high-quality puzzles over six high-level patterns across five distinct task configurations. In Table ~\ref{tab:dataset}, we compare \textsc{MARVEL} with existing AVR benchmarks to show its comprehensive scope.

\section{Hierarchical Evaluation Framework}
\label{sec: evaluation_framework}
Previous works evaluate MLLMs on AVR benchmarks with the final answer only~\citep{moskvichev2023conceptarc,mitchell2023comparing}, potentially overlooking shortcut learning and inductive biases~\citep{malkinski2023review}. On the other hand, precisely comprehending visual details is the foundation for subsequent reasoning in AVR problems~\citep{gao2023g}. We enrich \textsc{MARVEL} puzzles with perception questions~\citep{selvaraju2020squinting} designed to test models' perception ability on visual details~(\autoref{fig:example}). We design a hierarchical evaluation framework by combining two types of perception questions with AVR questions~(\Cref{fig:example}) to examine if model accuracy is based on perception and reasoning. For each puzzle, our framework provides three coarse-grained questions and one pattern-related fine-grained question:

\textbf{Coarse-grained Perception Question} in an open-ended fashion aims to test if models can understand task configuration correctly, which directly asks about the number of panels in puzzles. We use templates to generate three questions focusing on the number of panels in the context part, choice parts, and the whole puzzle. We remove the choice index when testing models with this question to avoid shortcut learning.

\textbf{Fine-grained Perception Question} in binary-choice format examines models' understanding of input shapes, which focus on the visual details~\citep{selvaraju2020squinting} such as shape attributes ~(number of edges) and spatial relationship~(left, right) based on the pattern contained in the puzzle. For example, in \Cref{fig:example}, the fine-grained perception question tests whether models can understand the number of circles because the puzzle contains \textit{Mathematical Pattern}. For each puzzle, we randomly pick one choice panel in the puzzle and manually create questions with two choices. The correct answer is randomly placed to avoid inductive bias. 
We have five types of questions based on the pattern and how it adapts to the input shape, which are listed with examples here:
\begin{enumerate}[leftmargin=*,itemsep=1pt,topsep=0pt,parsep=0pt]
    \item \textbf{Location}: Is the dot outside or inside of the star in choice 4?
    \item \textbf{Color}: Is the triangle black or white in choice 1?
    \item \textbf{Shape}: Is there a circle or a triangle inside choice 3?
    \item \textbf{Quantity}: Are there five or four circles in choice 2?
    \item \textbf{Comparison}: Are the left and right halves of the rectangle in choice 3 the same?

\end{enumerate}

\section{Experimental Setup}

\subsection{Model Selection}
\paragraph{Closed-source MLLMs.} We include API-based MLLMs including 1)~GPT-4V~\citep{openai2023gpt4}, 2)~Gemini~\citep{geminiteam2023gemini} and 3)~Claude3~\citep{claude3team2023anthropic}. With the massive computation and training data, these models show promising performance on a wide range of visual-focused tasks~\citep{vqav2,xie2019visual}. We evaluate closed-source MLLMs in both zero-shot and few-shot~\citep{brown2020language} settings.

\paragraph{Open-source MLLMs.} We include MLLMs smaller than 13B due to our limited computing resources: 1)~InstructBLIP~\citep{dai2024instructblip}, 2)~BLIP-2~\citep{li2023blip}, 3)~Fuyu~\citep{bavishi2023fuyu}, 4)~Qwen-VL~\citep{bai2023qwen} and 5)~LLaVA~\citep{liu2023visual}. We only evaluate these MLLMs in a zero-shot setting due to their single-image input settings~\citep{zhao2023mmicl}.

\paragraph{Human Evaluation.} To access the upper bound performance on \textsc{MARVEL}, we simulate a realistic human assessment by inviting 30 annotators aged from 10 to 50 years to solve a subset of \textsc{MARVEL} and ensure each subset contains every pattern in all task configurations. We compute the average performance of these 30 annotators as the human baseline. Each puzzle is solved by at least two annotators.

\subsection{Evaluation Metrics} Following a similar setting as previous research evaluating MLLMs on the AVR benchmark~\citep{ahrabian2024curious}, we use regex matching to extract the choices picked (e.g., "choice 4" in the response "The correct answer is choice 4."), with failure cases re-extracted by GPT-4~\citep{aher2023using}. We use accuracy as the metric, which is commonly used for evaluating multiple-choice questions and has been utilized by many AVR systems~\citep{zhang2019raven,hill2019learning}. 
Based on the hierarchical evaluation framework,
we evaluate MLLMs with two types of accuracy-based metrics: 

\begin{enumerate}[leftmargin=*,itemsep=1pt,topsep=-5pt,parsep=0pt]
    \item \textbf{Instance-based Accuracy} considers questions separately. We report accuracy results for \textit{AVR question} and \textit{fine-grained perception question}. 
    \item \textbf{Group-based Accuracy} considers questions as group to assess the consistency in model reasoning~\citep{jiang2023brainteaser,yuan2021perception}. The model receives a score of 1 only if it correctly answers all questions within the same group. We report the group-based accuracy result of combining all three coarse-grained perception questions and the further result after introducing fine-grained and AVR questions into the group.
\end{enumerate}

\begin{table}[h]
\renewcommand{\arraystretch}{0.95}
\centering
\resizebox{\textwidth}{!}{%
\begin{tabular}{lc|c|c|c|c|c}
\toprule
\multicolumn{1}{l}{\multirow{2}{*}{\textbf{Category}}} & \multirow{2}{*}{\textbf{Model}} & \multirow{1}{*}{\textbf{AVR}} & \multicolumn{1}{c|}{\textbf{Fine-grained}} & \multirow{2}{*}{\textbf{$\textbf{Perc}^\textbf{C}$}} &  \multirow{2}{*}{\textbf{$\textbf{Perc}^\textbf{C\&F}$}} & \multirow{2}{*}{\textbf{$\textbf{Perc}^\textbf{C\&F}$} \& \textbf{AVR}} \\
& & \multirow{1}{*}{\textbf{Question}} & \multicolumn{1}{c|}{\textbf{Perception}} &  & \\
\midrule
\multicolumn{2}{c|}{\textbf{Random}} & 25.00 & 50.00 & - & - & - \\
\midrule
\multicolumn{1}{l}{\multirow{5}{*}{\begin{tabular}[c]{@{}l@{}}\textbf{Open-}\\ \textbf{source} \\ \textbf{MLLMs}\end{tabular}}}
& \multicolumn{1}{l|}{\texttt{Qwen-VL (7B)}} & 19.61 & 37.27 & \underline{0.52} & 0.39 & 0.00 \\
& \multicolumn{1}{l|}{\texttt{Fuyu (8B)}} & 24.29$^\dag$ & 34.94 & 0.00 & 0.00 & 0.00 \\
& \multicolumn{1}{l|}{\texttt{BLIP-2 (FlanT5$_{\mathrm{XXL}}$-11B)}} & 24.81$^\dag$ & \underline{\textbf{53.38}} & 1.04 & \underline{0.52} & \underline{0.26} \\
& \multicolumn{1}{l|}{\texttt{InstructBLIP (Vicuna-13B)}} & 24.68$^\dag$ & 49.48 &  0.00 & 0.00 & 0.00 \\
& \multicolumn{1}{l|}{\texttt{LLaVA-1.5 (Vicuna-13B)}} & \underline{26.36} & 51.43 & \underline{1.14} & \underline{0.52} & 0.13 \\
\midrule
\multicolumn{1}{l}{\multirow{4}{*}{\begin{tabular}[c]{@{}l@{}}\textbf{Closed-} \\ \textbf{source} \\ \textbf{MLLMs}\end{tabular}}}
& \multicolumn{1}{l|}{\texttt{GPT-4V}} & 22.34 & \underline{51.56} & 18.31 & 9.22 & 2.73 \\
& \multicolumn{1}{l|}{\texttt{Gemini-pro-vision}*} & 25.06$^\dag$& 44.42 & 15.19 & 6.75 & 1.69 \\
& \multicolumn{1}{l|}{\texttt{Claude3~(Sonnet)}} & 26.49$^\dag$ & 50.91 & 38.70 & 19.87 & 5.06 \\
& \multicolumn{1}{l|}{\texttt{Claude3~(Opus)}} & \underline{\textbf{28.83}} & 47.27 & \underline{\textbf{44.94}} &\underline{\textbf{20.13}} & \underline{\textbf{5.97}} \\
\midrule
\multicolumn{2}{c|}{\textbf{Human}} & 68.86 $\pm$ 9.74 & - & - & -  & -  \\
\bottomrule
\end{tabular}%
}
\caption{Main zero-shot accuracy over \textsc{MARVEL} across all MLLMs in two accuracy metrics: $Prec^C$ = group-based accuracy over all coarse-grained perception questions (model must answer all three questions correctly), $Perc^{C\&F}$ = group-based accuracy combining all perception questions (coarse/fine-grained), AVR = AVR Question. The best performance among all models is in \textbf{bold}, and the best result in two MLLMs categories is \underline{underlined}. *Gemini refuses to answer the puzzle due to safety problems in 7\% cases so the performance is computed based on the left set. $\dag$ notes the result is attributed to inductive bias.\\}
\label{tab:main}
\vspace{-1.0cm}
\end{table}

\section{Results}

We focus on four research questions: 1) What's the abstract reasoning ability on visual puzzles of current SOTA MLLMs? 2)~Can MLLMs do better with different few-shot prompting strategies? 3)~How do MLLMs perform on different patterns and task configurations? 4)~To what extent do MLLMs visually understand the puzzle, and do they show 
consistent reasoning ability?

\textbf{Overall Performance.} The AVR question results are shown in~\Cref{tab:main}. Human performance reaches 68.86\%, with a standard deviation of 9.74, confirming the validity and challenging nature of \textsc{MARVEL}. For both open and closed source categories, \textbf{all models show near-random performance with a huge gap~(40\%) compared to human performance}, in which closed-source MLLMs~(avg: 25.7\%) perform slightly better than open-sourced ones~(avg: 24.0\%). We observed an extremely imbalanced distribution in the outputs of some MLLMs. For example, BLIP2 consistently selecting choice 1 for all puzzles~(marked $\dag$ in ~\Cref{tab:main}). We tried different approaches with 
\begin{wrapfigure}{r}{0.35\textwidth}
{
  \includegraphics[width=\linewidth]{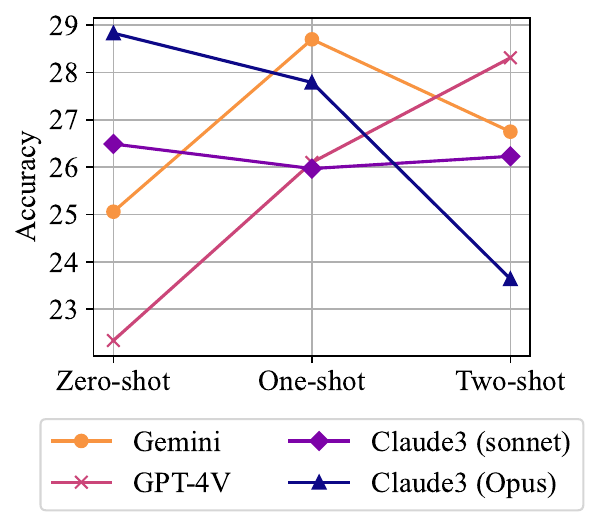}
  \caption{MLLMs performance in different few-shot COT.}
  \label{fig: prompting}
  \vspace{-0.3cm}
}
\end{wrapfigure}
our best effort to avoid potential bad prompts or engineering settings, including adding question marks in the black panel, replacing the choice index with letter~(1 $\rightarrow$ A), and changing the description in the AVR question. None of them can mitigate and may even exacerbate the issue, highlighting the potential inductive biases~\citep{wang2024theoretical} in models. Among open-source MLLMs, LLAVA performs the best, yet the gap is very small, and it is unclear whether the gain comes from its larger model size. 
In closed-source models, even the strongest MLLM, Claude3~(Opus), which demonstrated promising results on various vision tasks~\citep{claude3team2023anthropic}, failed to present a significant performance difference from the random baseline. Claude3~(Sonnet) and Gemini also have imbalanced output distributions, with both selecting choice 4 in most cases.


\textbf{Impact of Few-Shot CoT.}
Given the poor zero-shot performance and models' capability of in-context learning~\citep{brown2020language}, we explore few-shot prompting with Chain-of-Thought~(CoT) \citep{wei2022chain} to guide MLLMs with abstract reasoning patterns.
We experiment with all closed-source MLLMs in one-shot and two-shot settings using manually created CoT context, similar to~\citet{yang2023dawn}. For each puzzle, we randomly select puzzles with the same pattern in the sequence task configuration and annotate the CoT reasoning with answers as demonstrations. We chose the sequence task because it is more straightforward~(only along the sequence) compared to other configurations. Each demonstration is formatted as image-text pairs. Our result is shown in~\Cref{fig: prompting}, and we present the full results in~\Cref{app: fewshot}. The few-shot demonstrations show a marginal positive impact on GPT-4V and a decreasing trend on Claude3~(Opus).

\begin{wrapfigure}{r}{0.50\textwidth}
{
  \centering
  \vspace{-0.5cm}
  \includegraphics[width=\linewidth]{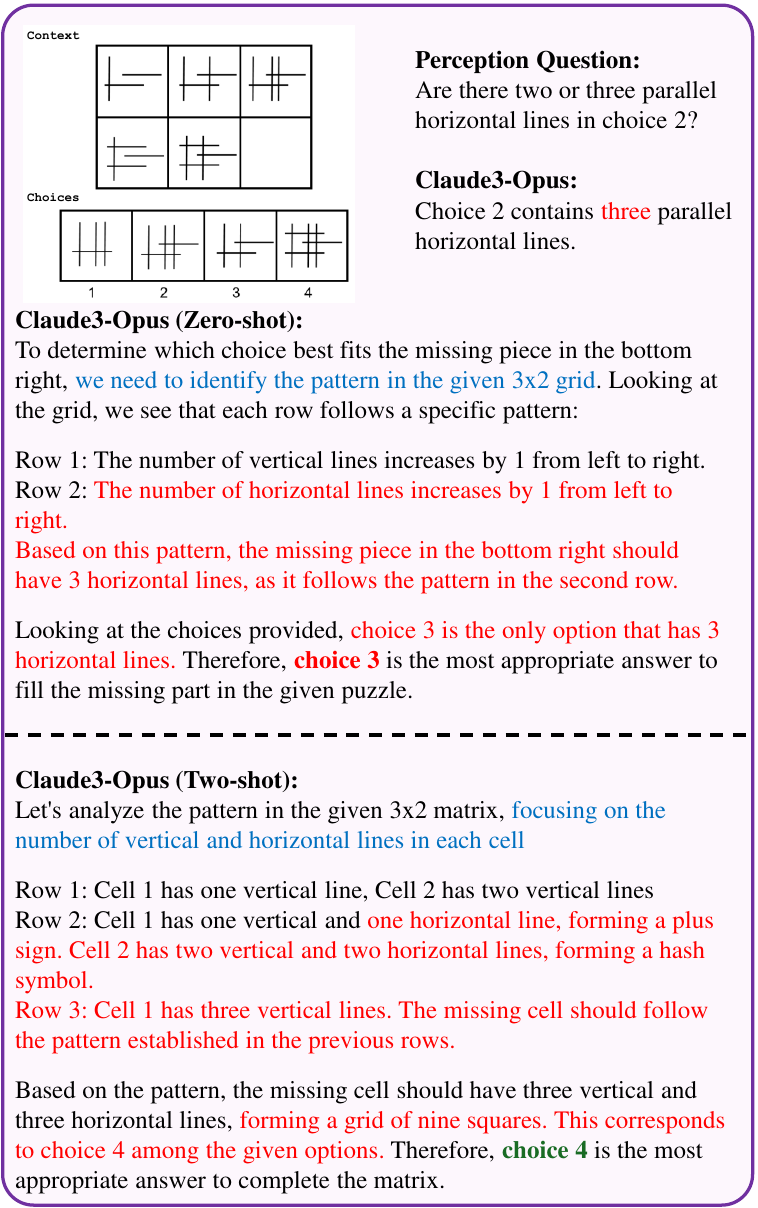}
  \caption{Perception question, zero- and two-shot example of Claude3~(Opus).}
      \label{fig:few-shot example}
  \vspace{-0.4cm}
}
\end{wrapfigure}

Further analysis reveals that the main improvement in GPT-4V's results lies in the \textit{3D-Geometry pattern}. As this pattern focuses on reassembling, the demonstration can guide the model to pay attention to the relative position of each face of the object. However, since most patterns are uniquely implemented on different input shapes and their attributes, the model struggles to learn generalizable patterns from the few-shot demonstrations. \Cref{fig:few-shot example} provides an example of zero-and few-shot results from Claude3~(Opus). With the demonstration, the model learns to focus on the correct pattern~(blue) at the beginning of the reasoning. However, it fails to adapt precisely to the input shapes in the puzzle~(red), leading to errors in subsequent reasoning. We also test different prompt engineering approaches, including selecting demonstration samples from different 1)~patterns, 2)~task configurations, and 3)~prompting MLLMs by dividing puzzles panel by panel. None of these approaches yields a positive impact; instead, they lead to a significant drop in performance~(Appendix\ref{app: fewshot}). Given the complexity and challenging nature of the dataset, the effectiveness of few-shot prompting on \textsc{MARVEL} remains minimal.\footnote{We find the inductive bias is mitigated in few-shot settings.}

\begin{figure*}[t]
  \centering
  \vspace{-0.5cm}
  \includegraphics[trim=0 0 0 0, clip, width=0.76\textwidth]{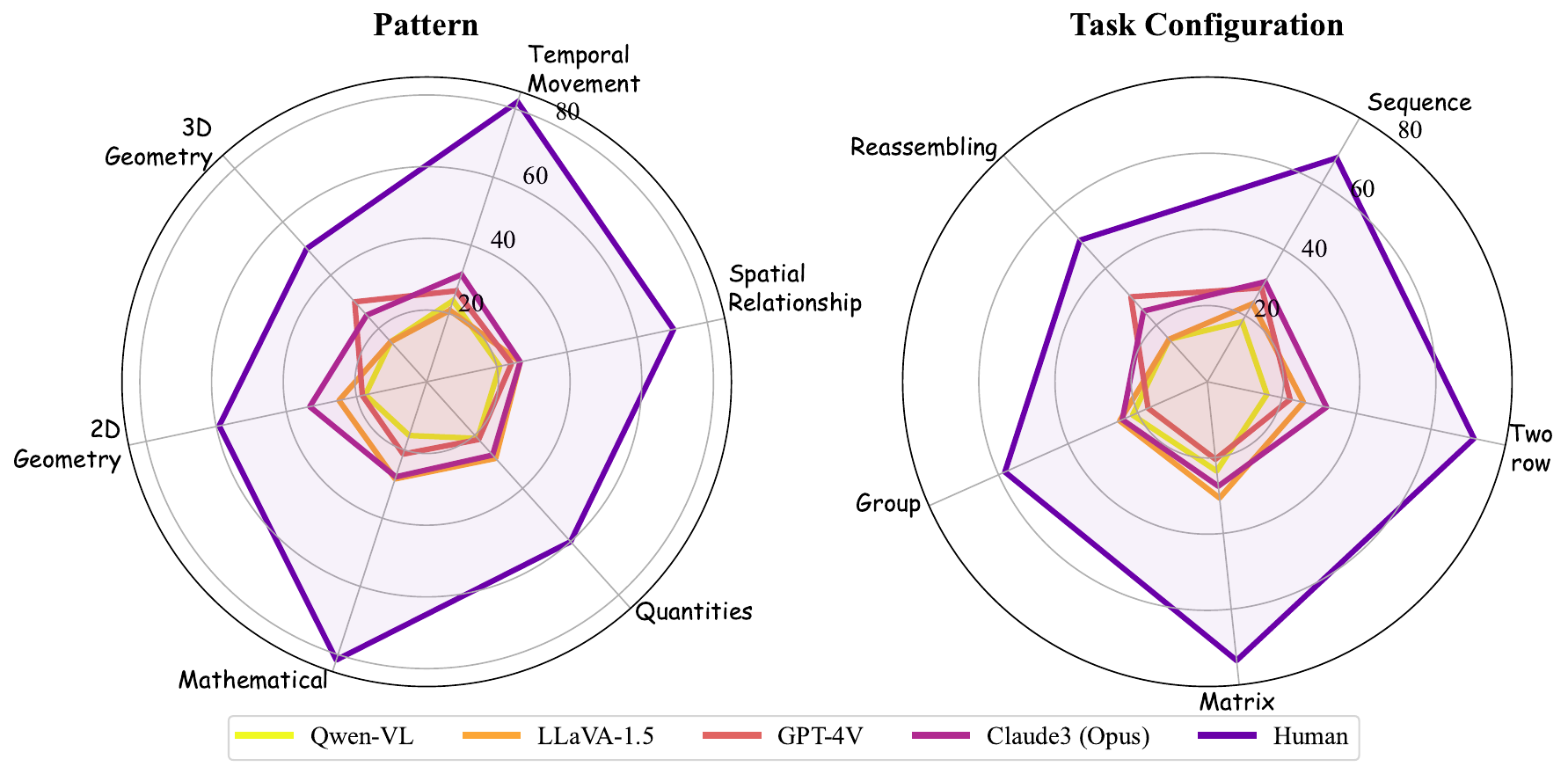}
  \caption{
    MLLMs and human performance across patterns and task configurations.
  }
  \label{fig:zsperformance}
  \vspace{-0.5cm}
\end{figure*}

\textbf{Performance on Different Patterns and Configurations.} 
We further break down the results\footnote{We only select MLLMs that do not show significantly imbalanced output distributions.} into different patterns and task configurations in~\Cref{fig:zsperformance}~(full results in \Cref{app: concept and tasks}). In general, MLLMs show near-random performance on all patterns and task configurations. Among the six patterns, \textit{3D-Geomertry} pattern is the most challenging for humans and MLLMs. The difficulty may be rooted in the requirement for the 3D imaginative ability~\citep{margules1988heading} and relatively less common in the models' pre-training datasets, which leads to a significant gap~(12.5\%) between open and closed-sources MLLMs. On the other hand, \textit{2D-Geomertry}, involving understanding geometric attribute of input shape, seems relatively easier for MLLMs, evidenced by their proficiency on similar tasks such as interpreting graphs plots~\citep{yang2023dawn}. 
In two model categories, LLaVA ranks first on three patterns, showing competitive performance compared to closed-source MLLMs. Claude3~(Opus) exhibits a balanced and strong reasoning ability, ranking in the top 2 across all patterns.

Among the five task configurations, the two-row and matrix formats provide more panels (i.e., information) to verify the pattern, facilitating puzzle-solving. On the contrary, the group and reassembling formats containing the least panel number tend to be challenging. Three out of four MLLMs rank 1st in different task configurations, which verifies our assumption of potential bias in single-configuration evaluation. Models may be familiar with specific input types according to their pre-training dataset, highlighting the importance and necessity of multidimensional settings in \textsc{MARVEL}.

\textbf{Perception Ability and Reasoning Consistency.}
Visual cognition forms the foundation for advanced reasoning~\citep{richards1984parts}. By incorporating perception questions, our hierarchical evaluation framework effectively 
investigates to what extent the models understand the visual information from the puzzle. In \Cref{tab:main}, closed-source MLLMs demonstrate more robust performance on coarse-grained perception group accuracy compared to open-sourced MLLMs, with a performance gap ranging from 14.05\% to 44.80\%. However, even the best model fails to reach 50\% accuracy, indicating that current MLLMs struggle to simultaneously understand the number of grids, choices, and the puzzle as a whole, despite their promising performance on real-world datasets~\citep{hudson2019gqa}. The simplicity of the coarse-grained perception questions (all puzzles contain less than 13 panels) highlights the poor perception ability of current MLLMs in the abstract visual reasoning domain. Fine-grained perception questions further confirm this argument, with all models showing near-random performance. Further analysis of fine-grained perception performance based on five categories~({Table in \Cref{app: perception}}) reveals that models perform relatively better at color perception but have difficulty recognizing location~(e.g., `a' is on the left of `b'). We hypothesize that the difficulty in understanding location stems from the lack of labeled data on location and relations during training, especially in abstract visual understanding. In contrast, the models' color perception is well-trained during their multi-modal alignment, and the simplicity of RGB understanding allows for easier transfer to the abstract domain.

The further group-based accuracy~($Prec^{C\&F}$ and $Prec^{C\&F} \& AVR$) shows that~\textbf{no model can solve the AVR puzzles with consistent reasoning}, with the best model reaching only 5.97\% group accuracy. Based on the result of our evaluation framework, we hypothesize the inconsistency stems from their poor visual perception ability~\citep{gao2023g}. To verify our assumption, we run an additional experiment on a subset of \textsc{MARVEL} by adding accurate text descriptions of the puzzle in the input~(full result in \Cref{app: text}). The result shows a significant boost in performance~(11.57\% to 44.21\%), with GPT-4V even displays on-par
performance (65.26\%) with humans. As shown in
\Cref{fig:few-shot example}, the model's reasoning is based on the perception of the puzzle~(e.g., number of lines), which needs to be completely precise to support correct reasoning.  The perception questions in our framework reveal that the model cannot clearly understand the number of lines, explaining why it fails to answer the puzzle even with correct hints (few-shot). A single error in visual feature perception can impact reasoning since the correct pattern must apply to all puzzle shapes. The densely packed information distribution—where the majority of the puzzle remains blank—ensures that each piece of visual perception is an essential foundation for subsequent reasoning. However, the importance of visual detail perception has received little attention in previous evaluations~\citep{moskvichev2023conceptarc,mitchell2023comparing}, highlighting the significance of our new evaluation benchmark.

\section{Conclusion}

In this work, we developed \textsc{MARVEL}, a multidimensional abstract reasoning benchmark comprising 770 puzzles with both geometric and abstract input shapes across six patterns and five task configurations. We also designed a hierarchical evaluation framework that enriches MARVEL with perception questions to enable granular analysis of models' visual details understanding and reasoning consistency. Our comprehensive experiments with nine SoTA MLLMs revealed a huge gap in abstract visual reasoning ability between~(40\%) humans and MLLMs, where all MLLM often perform close to random. Further analysis based on our evaluation framework showed MLLMs' poor perception ability in understanding visual details, which hinders their subsequent reasoning and leads to poor AVR performance. We hope future works can build on the foundation of \textsc{MARVEL} for enhancing MLLM abstract visual perception and reasoning abilities.
\section{Acknowledgements} We appreciate Fred Morstatter for very helpful comments. We thank Tian Jin and Jiachi Liang for their assistance in data collection. This research was sponsored by the Defense Advanced Research Projects Agency via Contract HR00112390061.

\bibliography{cite}
\bibliographystyle{colm2024_conference}

\appendix
\section{Data Examples}
\label{app: examples}

\begin{figure}[htbp]
  \centering
\includegraphics[width=0.6\linewidth]{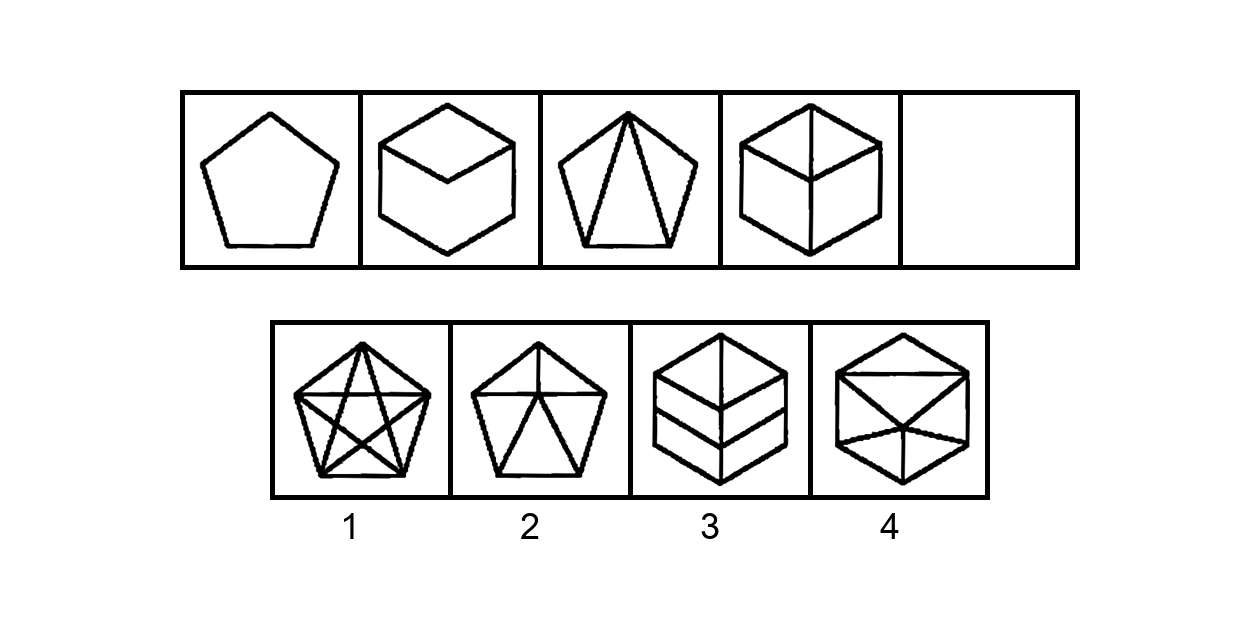}
  \caption{
  The example is formatted in \textit{Sequence} configuration with the \textit{Quantities} pattern. The answer to this puzzle is B.
  }
  \label{fig:example0}
\end{figure}

\begin{figure}[htbp]  
  \centering
  \includegraphics[width=0.6\textwidth]{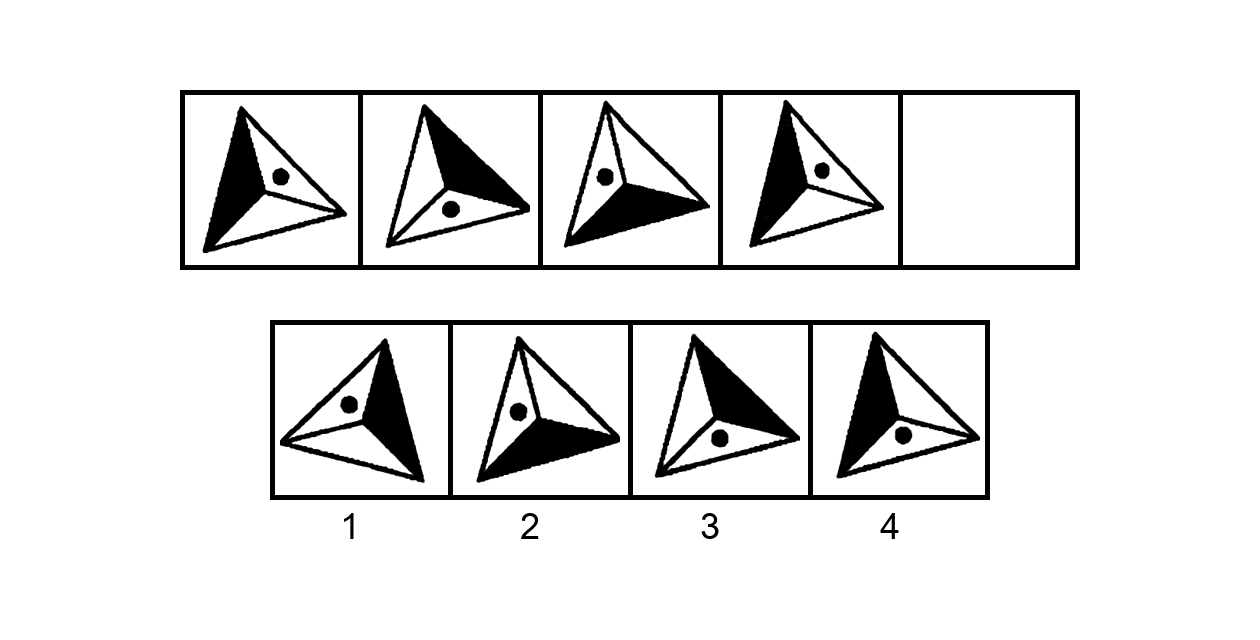}
  \caption{
    The example is formatted in \textit{Sequence} configuration with the \textit{Temporal Movement} pattern. The answer to this puzzle is C.
  }
  \label{fig:example1}
\end{figure}

\begin{figure}[htbp]
  \centering
  \includegraphics[width=0.6\textwidth]{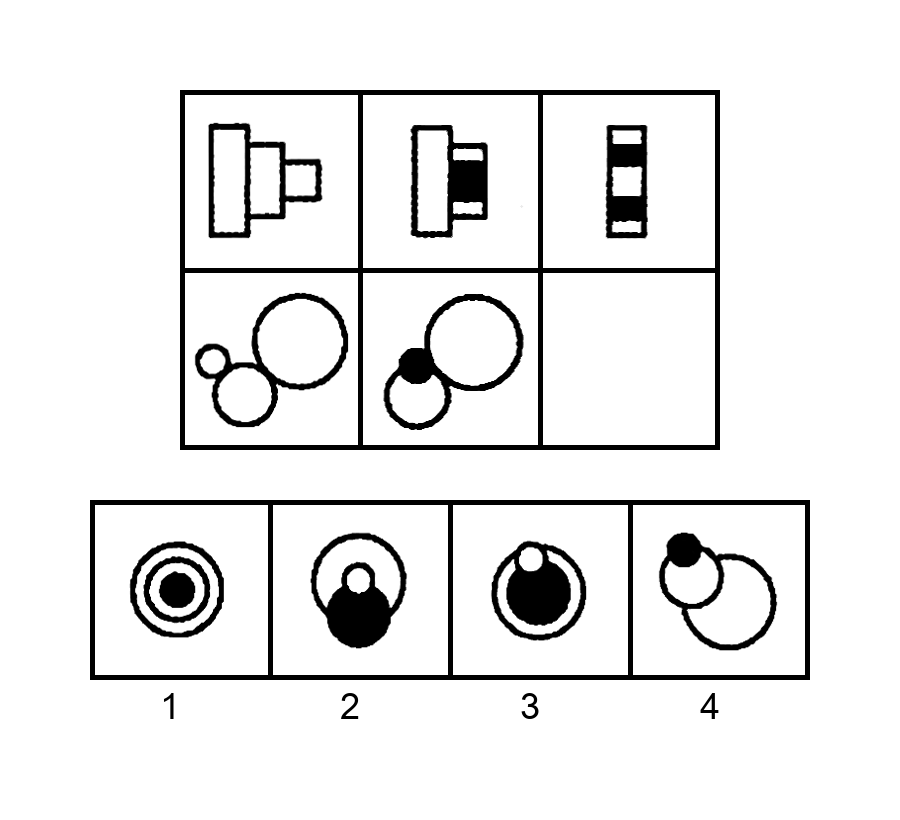}
  \caption{
    The example is formatted in \textit{Two-row} configuration with the \textit{Spatial Relationship} pattern. The answer to this puzzle is B.
  }
  \label{fig:example2}
\end{figure}

\begin{figure}[htbp]
  \centering
  \includegraphics[width=0.6\textwidth]{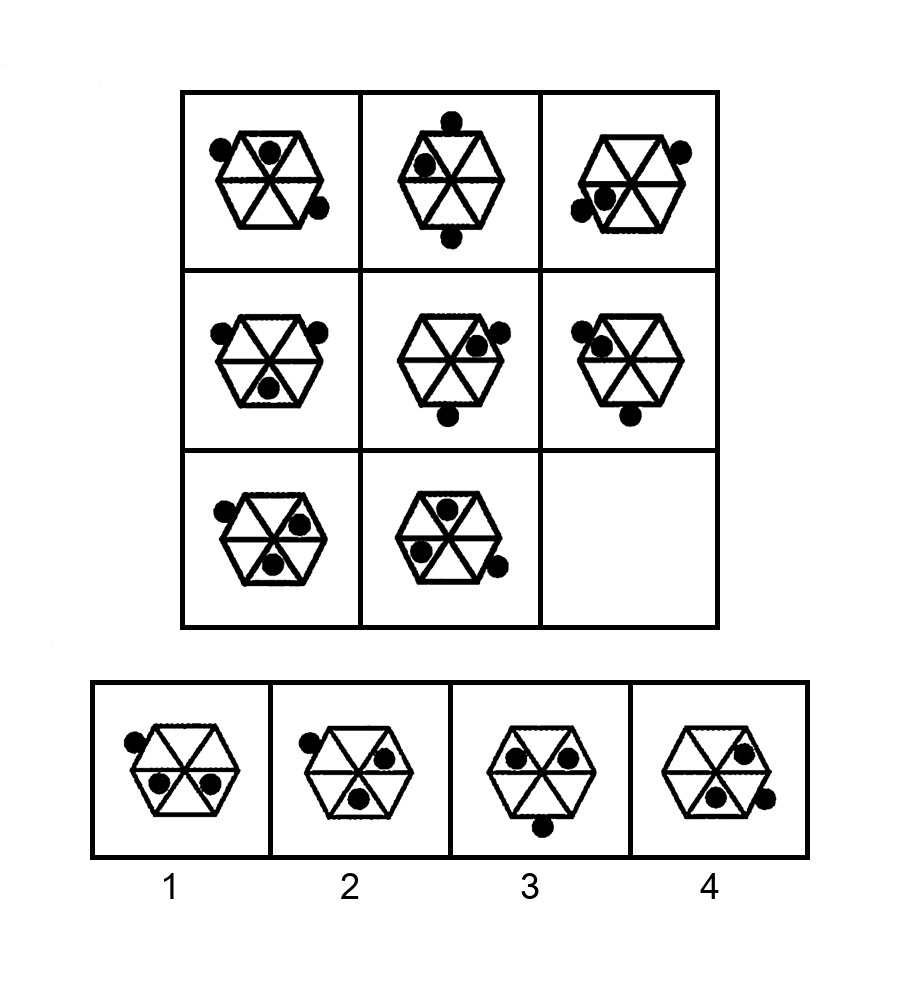}
  \caption{
    The example is formatted in \textit{Matrix} configuration with the \textit{Spatial Relationship} pattern. The answer to this puzzle is B.
  }
  \label{fig:example3}
\end{figure}

\begin{figure}[htbp]
  \centering
  \includegraphics[width=0.6\textwidth]{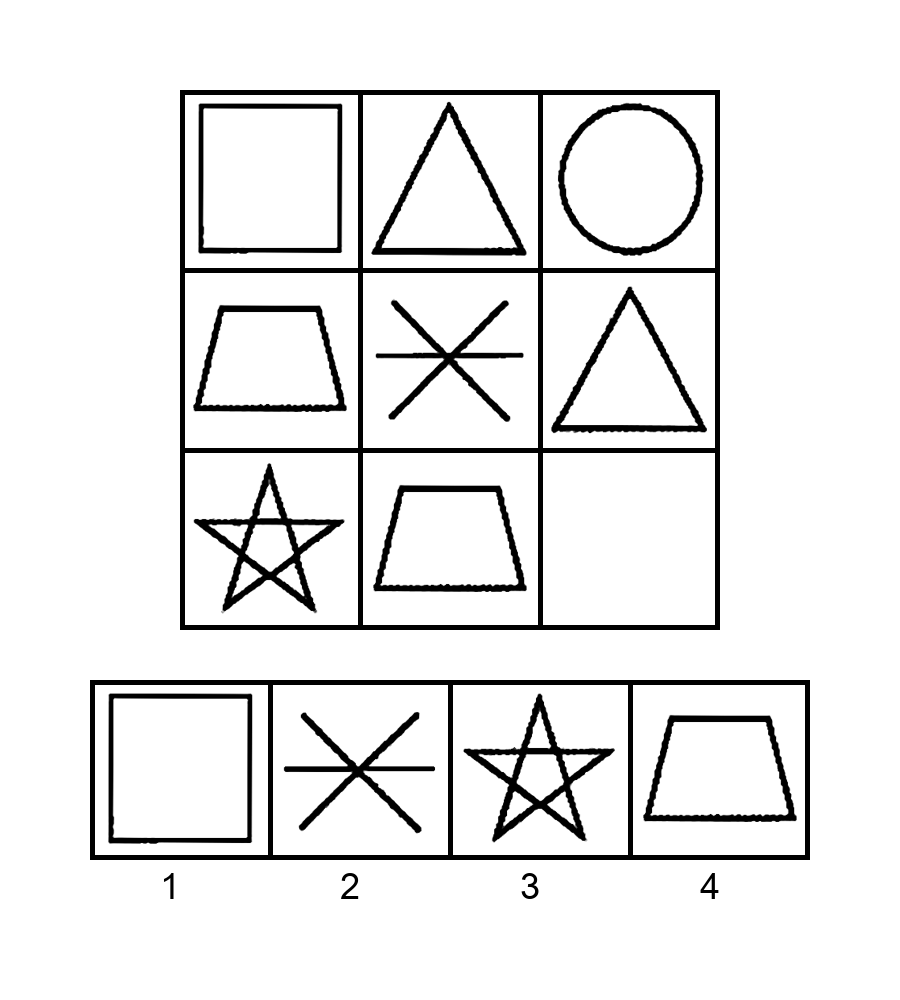}
  \caption{
    The example is formatted in \textit{Matrix} configuration with the \textit{Mathematical} pattern in \textit{Geometric} shapes. The answer to this puzzle is A.
  }
  \label{fig:example4}
\end{figure}

\begin{figure}[h]
  \centering
\includegraphics[width=0.6\textwidth]{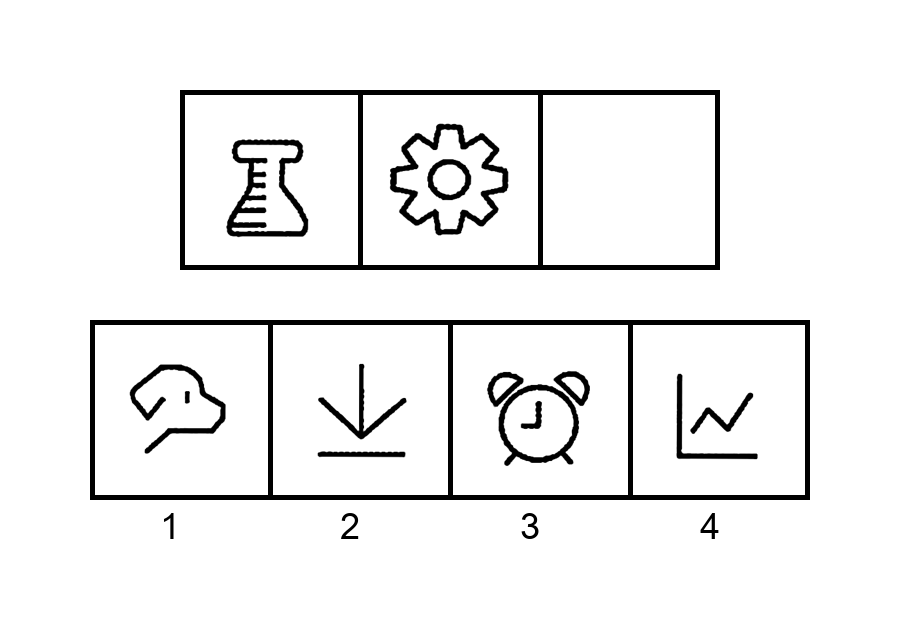}
  \caption{
    The example is formatted in \textit{Group} configuration with the \textit{2D-Geometry} pattern in \textit{Abstract} input shapes. The answer to this puzzle is C.
  }
  \label{fig:example5}
\end{figure}

\begin{figure}[h]
  \centering
  \includegraphics[width=0.6\textwidth]{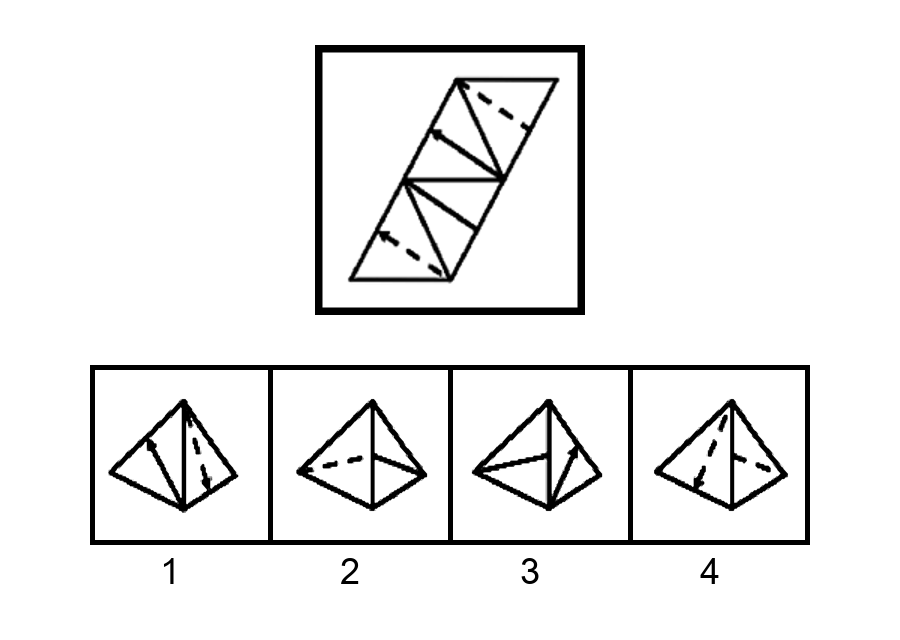}
  \caption{
    The example is formatted in \textit{Reassembling} configuration with the \textit{3D-Geometry} Quantities. The answer to this puzzle is B.
  }
  \label{fig:example6}
\end{figure}

We present examples with 5 different task configurations and 6 patterns in \Cref{fig:example0,fig:example1,fig:example2,fig:example3,fig:example4,fig:example5,fig:example6}.


\begin{figure}[t!]
  \centering
  \includegraphics[width=0.8\textwidth]{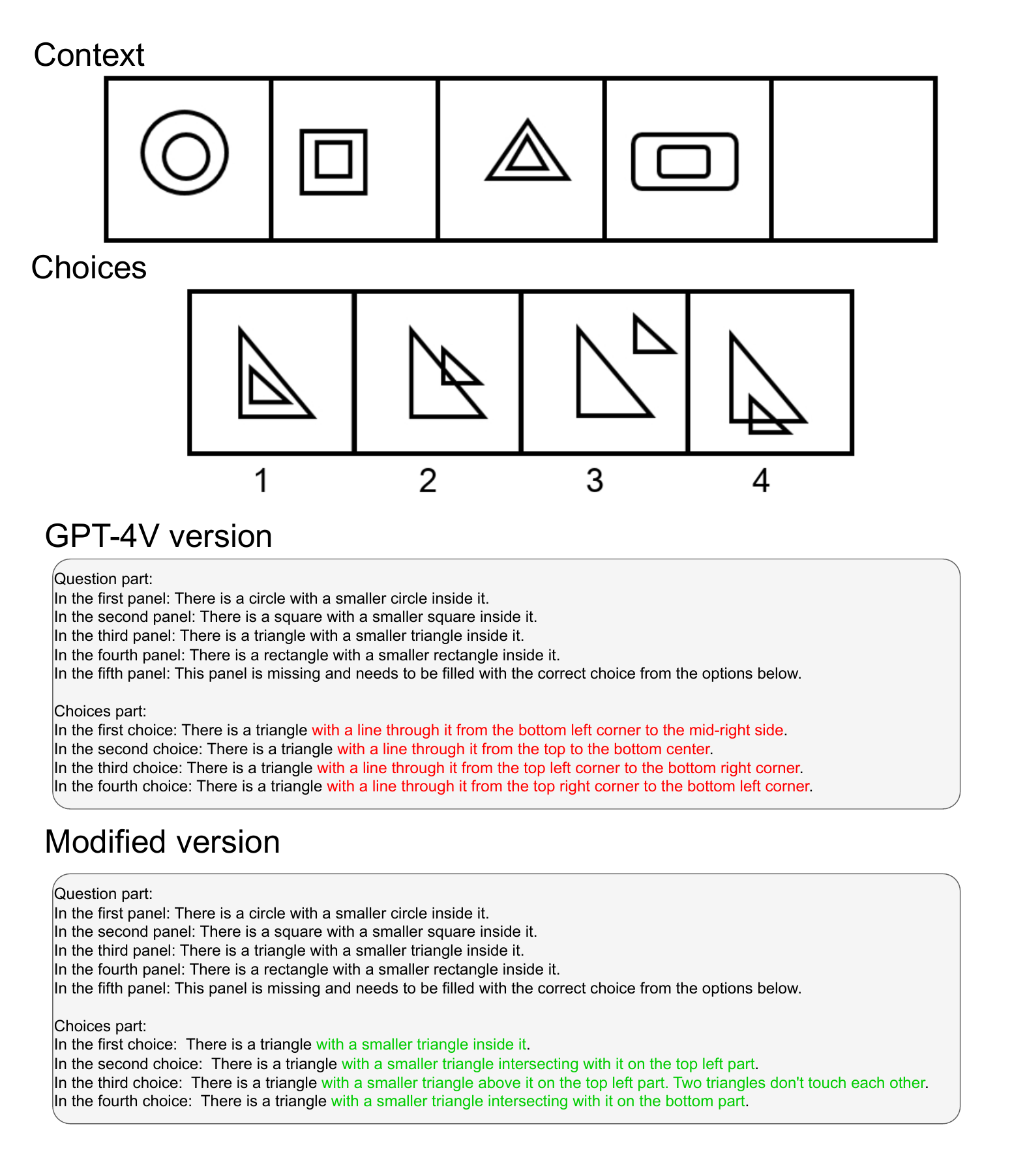}
  \caption{
 An example of the annotation process. We first generate text descriptions from GPT-4V, and the output will further be modified by human annotators~(\textcolor{red}{red} $\rightarrow$ \textcolor{green}{green}). 
  }
  \label{fig:text_detail}
\end{figure}
\section{Performance on Different Pattern and Tasks}
\label{app: concept and tasks}

\autoref{ablation_highlevel} and \autoref{ablation_task} shows MLLMs' AVR question performance on different patterns and tasks.

\begin{table}[h]
\centering
\resizebox{\textwidth}{!}{%
\begin{tabular}{l|c|c|c|c|c|c|c|c|c|c}
\toprule
\multirow{2}{*}{Pattern} & \multirow{2}{*}{Human} & \multicolumn{5}{c|}{Open-sourced MLLMs} & \multicolumn{4}{c}{Closed-sourced MLLMs} \\
\hhline{|~|~|---------}
 &   & Qwen-VL & Fuyu$\dag$ & Blip-2$\dag$ & InstructBLIP$\dag$ & LLaVA-1.5 & GPT-4V & Gemini$\dag$ & Claude3~(Sonnet)$\dag$ & Claude3~(Opus)\\
\cmidrule{1-11}
Temporal Movement & 82.08 & 23.81 & 24.76 & 25.71 & 25.71 & 20.95 & 26.67 & 22.86 & 23.81 & 31.43\\
Spatial Relationship & 70.42 & 20.83 & 25.00 & 26.67 & 26.67 & 26.67 & 24.17 & 29.17 & 34.17 & 26.67\\
Quantities & 81.57 & 15.76 & 24.85 & 27.88 & 27.27 & 28.48 & 21.21 & 24.85 & 24.85 & 27.88\\
Mathematical & 60.00 & 21.25 & 25.00 & 24.17 & 24.17 & 28.75 & 21.67 & 23.75 & 23.33 & 27.50\\
2D-Geometry & 59.17 & 17.50 & 22.50 & 20.83 & 20.83 & 25.00 & 18.33 & 25.00 & 30.83 & 33.33\\
3D-Geometry & 50.00 & 15.00 & 15.00 & 15.00 & 15.00 & 15.00 & 30.00 & 30.00 & 20.00 & 25.00\\
\bottomrule
\end{tabular}%
}
\caption{Performance of different models for different patterns. $\dag$ notes the result is attributed to inductive bias.}
\label{ablation_highlevel}
\end{table}

\begin{table}[h]
\centering
\resizebox{\textwidth}{!}{%
\begin{tabular}{l|c|c|c|c|c|c|c|c|c|c}
\toprule
\multirow{2}{*}{Task format} &\multirow{2}{*}{Human} & \multicolumn{5}{c|}{Open-sourced MLLMs} &  \multicolumn{4}{c}{Closed-sourced MLLMs}\\
\hhline{|~|~|---------}
 &   & Qwen-VL & Fuyu$\dag$ & Blip-2$\dag$ & InstructBLIP$\dag$ & LLaVA-1.5 & GPT-4V & Gemini$\dag$ & Claude3~(Sonnet)$\dag$ & Claude3~(Opus)\\
 \cmidrule{1-11}
Sequence & 67.84 & 18.18 & 22.42 & 23.03 & 23.03 & 23.64 & 28.48 & 23.03 & 26.06 & 30.30\\
Two-row & 71.56 & 16.00 & 22.67 & 24.89 & 24.44 & 25.78 & 22.22 & 24.00 & 25.78 & 32.00\\
Matrix & 73.56 & 23.56 & 29.78 & 28.89 & 28.89 & 30.67 & 20.44 & 26.67 & 27.56 & 27.56\\
Group & 58.18 & 21.48 & 21.48 & 21.48 & 21.48 & 25.19 & 17.04 & 25.93 & 27.41 & 24.44\\
Reassembling & 50.00 & 15.00 & 15.00 & 15.00 & 15.00 & 15.00 & 30.00 & 30.00 & 20.00 & 25.00 \\
\bottomrule
\end{tabular}%
}
\caption{Performance of different models for different task configurations. $\dag$ notes the result is attributed to inductive bias.}
\label{ablation_task}
\end{table}
\section{Coarse-grained Perception on Different Categories}
\label{app: perception}

\autoref{coarse-grained} shows MLLMs coarse-grained perception performance on different categories.

\begin{table}[t]
\centering
\resizebox{\textwidth}{!}{%
\begin{tabular}{l|c|c|c|c|c|c|c|c|c}
\toprule
Category & Qwen-VL & Fuyu & Blip2 & InstructBLIP & LLaVA-1.5 & GPT-4V & Gemini & Claude3~(Sonnet)* & Claude3~(Opus)\\
\midrule
Location & 40.00 & 15.38 & 40.00 & 35.38 & 61.54 & 41.54 & 41.54 & 32.31 & 32.31\\
Color & 37.74 & 26.42 & 58.49 & 58.49 & 58.49 & 52.83 & 54.72 & 47.17 & 62.26\\
Shape & 33.33 & 36.16 & 45.20 & 36.72 & 29.38 & 45.76 & 53.67 & 57.63 & 45.76\\
Quantity & 36.84 & 36.84 & 60.86 & 56.25 & 57.57 & 51.64 & 39.47 & 56.58 & 48.36\\
Comparison & 40.94 & 40.35 & 52.05 & 53.22 & 57.31 & 60.82 & 41.52 & 42.11 & 47.95\\
\bottomrule
\end{tabular}%
}
\caption{Coarse-grained Perception Performance on Different Categories
bias.}
\label{coarse-grained}
\end{table}

\section{Question with text description}
\label{app: text}
To investigate how the models perform subsequent reasoning with accurate visual detail awareness, we randomly select 95 puzzles~(five for each pattern in each task configuration) and provide text descriptions for each panel in the puzzle. To ensure a similar level of granularity for text descriptions, we first use GPT-4V to provide the original text descriptions, which human annotators further modify without introducing unrelated details~(\Cref{fig:text_detail}).

We input MLLMs with both the AVR question and text descriptions. The result is shown in 
\Cref{tab:text_detail}. With the help of text descriptions, MLLMs, especially closed-sourced MLLMs, can build their abstract visual reasoning on correct visual detail foundations, gaining significant improvement in the performance~(11.57\% to 44.21\%). GPT-4V even shows on-par performance~(65.26\%) with humans, highlighting the importance of visual perception ability. We also want to point out that some puzzles containing abstract shapes are challenging to describe. A tool that can convert images to SVGs and text descriptions~\citep{wang2024text} can be a possible approach to mitigate the difficulty and enhance MLLMs performance. 

\begin{table}[t]
\centering
\resizebox{\textwidth}{!}{%
\begin{tabular}{l|c|c|c|c|c|c|c|c|c}
\toprule
\multirow{2}{*}{Input} & \multicolumn{5}{c|}{Open-sourced MLLMs} & \multicolumn{4}{c}{Closed-sourced MLLMs} \\
\hhline{|~|---------}
    & Qwen-VL & Fuyu & Blip-2& InstructBLIP & LLaVA-1.5 & GPT-4V & Gemini & Claude3~(Sonnet)& Claude3~(Opus)\\
\cmidrule{1-10}
AVR  & 23.16&	23.16$\dag$	&23.16$\dag$	&23.16$\dag$	&21.05&	21.05	&26.32$\dag$	&27.37$\dag$&	30.53\\
AVR+Text  & 24.21&	26.32&	26.32	&13.68	&28.42	&65.26	&37.89&	49.47	&55.79\\
\bottomrule
\end{tabular}%
}
\caption{Performance of different models after introducing text description in the input. $\dag$ notes the result is attributed to inductive bias.}
\label{tab:text_detail}
\end{table}

\section{Few-Shot COT}
\label{app: fewshot}
We show our prompting template in \Cref{tab:prompt_cot}.
We discuss few-shot performance in \Cref{tab:few_shot_cot,tab:few_shot_cot_acc,few_shot_cot_ood,tab:fewshot_concept}. \Cref{tab:few_shot_cot_acc} shows few-shot result with Chain-of-Thought demonstrations in the prompt. \Cref{few_shot_cot_ood} compares one-shot result with Chain-of-Thought demonstrations from same pattern or different pattern picked randomly~(OOD). \Cref{tab:few_shot_cot} compares one-shot result with Chain-of-Thought demonstrations from single task format~(sequence) and two different task formats~(sequence and two-row).
\Cref{tab:fewshot_concept} shows the model's few-shot performance on different patterns.
\begin{table}[h]
\centering
\begin{tabular}{l|c|c|c}
\toprule
Model & zero-shot & one-shot & two-shot \\
\midrule
Gemini & 25.06 & 28.7 & 26.75 \\
GPT-4V & 22.34 & 26.1 & 28.31 \\
Claude3 (sonnet) & 26.49 & 25.97 & 26.23 \\
Claude3 (Opus) & 28.83 & 27.79 & 23.64 \\
\bottomrule
\end{tabular}%
\caption{Few-shot COT Accuracy}
\label{tab:few_shot_cot_acc}
\end{table}

\begin{table}[h]
\centering
\begin{tabular}{l|c|c}
\toprule
Model & one-shot & one-shot (OOD) \\
\midrule
Gemini & 23.85 & 24.62 \\
GPT-4V & 27.31 & 23.85 \\
Claude3 (sonnet) & 25.00 & 26.92 \\
Claude3 (Opus) & 29.62 & 25.00 \\
\bottomrule
\end{tabular}%
\caption{Few-shot COT Ablation}
\label{few_shot_cot_ood}
\end{table}

\begin{table}[h]
\centering
\begin{tabular}{l|c|c}
\toprule
Model & two-shot  & two-shot (mix) \\
\midrule
Gemini & 30.48 & 24.29 \\
GPT-4V & 29.52 & 28.10 \\
Claude3 (sonnet) & 28.57 & 23.81 \\
Claude3 (Opus) & 21.90 & 22.86 \\
\bottomrule
\end{tabular}%
\caption{Few-shot COT Ablation}
\label{tab:few_shot_cot}
\end{table}

\begin{table}[h]
\centering
\resizebox{\textwidth}{!}{
\begin{tabular}{l|c|c|c|c|c|c|c|c}
\toprule
 & \multicolumn{4}{c|}{one -shot}  & \multicolumn{4}{c}{two -shot} \\
 \midrule
 & Gemini & GPT-4V & Claude3 (Sonnet)  & Claude3 (Opus) & Gemini & GPT-4V & Claude3 (Sonnet)  & Claude3 (Opus)\\
 \midrule
Temporal Movement & 24.76 & 27.62 & 17.14 & 29.52 & 33.33 & 28.57 & 24.76 & 18.10\\
Spatial Relationship & 30.00 & 33.33 & 24.17 & 26.67 & 25.83 & 22.50 & 30.83 & 19.17\\
Quantities  & 21.21 & 27.88 & 27.88 & 26.67 & 25.45 & 31.52 & 25.45 & 27.27\\
Mathematical  & 27.92 & 28.75 & 25.42 & 26.67 & 25.83 & 27.50 & 22.92 & 22.08\\
2D-Geometry & 26.67 & 25.00 & 30.83 & 30.83 & 26.67 & 30.00 & 28.33 & 31.67\\
3D-Geometry  & 25.00 & 35.00 & 45.00 & 30.00 & 20.00 & 35.00 & 40.00 & 20.00\\
\bottomrule
\end{tabular}%
}
\caption{Few-shot performance on different patterns}
\label{tab:fewshot_concept}
\end{table}

\begin{table*}[t!]
\scriptsize
\centering
    \begin{tabularx}{\textwidth}{l|X}
        \toprule
        \textbf{Experiment} & \textbf{Prompt Example} \\ \midrule
        AVR Question (Reassembling) & \textit{[IMG]} You are given a puzzle. The puzzle consists of a context part on the top and the choices in the bottom. The context part on the top is an unfolded diagram of the 3D shape. The choices part on the bottom contains 4 options (marked by 1, 2, 3, or 4) represents the correct three-dimensional assembly. Which option (either 1, 2, 3, or 4) is the most appropriate answer? \\ \midrule

        AVR Question (Group) & \textit{[IMG]} You are given a puzzle. The puzzle consists of a context part on the top and the choices at the bottom. The context part on the top is a set of visual patterns arranged in a three-by-one sequence, with the last piece missing. The choices part on the bottom contains four options (marked by 1, 2, 3, or 4). Which option (1, 2, 3, or 4) is the most appropriate answer to fill the missing part? \\ \midrule

        AVR Question (Matrix) & \textit{[IMG]} You are given a puzzle. The puzzle consists of a context part on the top and the choices at the bottom. The context part on the top is a set of visual patterns arranged in a three by three matrix, with the bottom right piece missing. The choices part on the bottom contains four options (marked by 1, 2, 3, or 4). Which option (1, 2, 3, or 4) is the most appropriate answer to fill the missing part? \\ \midrule

        AVR Question (Sequence) & \textit{[IMG]} You are given a puzzle. The puzzle consists of a context part on the top and the choices part on the bottom. The context part on the top is a set of visual patterns arranged sequentially, with the last piece missing. The choices part on the bottom contains four options (marked by 1, 2, 3, or 4). Which option (1, 2, 3, or 4) is the most appropriate answer to fill the missing part? \\ \midrule
        
        AVR Question (Two-row) & \textit{[IMG]} You are given a puzzle. The puzzle consists of a context part on the top and the choices part on the bottom. The context part on the top is a set of visual patterns arranged in a three by two matrix, with the bottom right piece missing. The choices part on the bottom contains four options (marked by 1, 2, 3, or 4). Which option (1, 2, 3, or 4) is the most appropriate answer to fill the missing part? \\ \midrule

        Perception Question (fine-grained) & \textit{[IMG]} You are given a puzzle. The puzzle consists of a context part on the top and the choices part on the bottom. The context part on the top is a set of visual grids arranged in a \textit{{m} by {n}} sequence, with the last piece missing. The choices part on the bottom contains four choices (marked by 1, 2, 3, or 4). \textit{[question]} \\ \midrule

        Perception Question (coarse-grained) & \textit{[IMG]} You are given a puzzle. The puzzle consists of a context part on the top and the choices part on the bottom. The context part on the top contains some grids, with the last missing blank grid to be completed. The choices part on the bottom contains a sequence of grids representing the possible choices. How many grids, including the blank grid, are in the context part?  \\ \midrule

        Perception Question (coarse-grained) & \textit{[IMG]} You are given a puzzle. The puzzle consists of a question part on the top and the choices part on the bottom. The question part on the top contains some grids, with the last missing blank grid to be completed. The choices part on the bottom contains a sequence of grids representing the possible choices. How many grids are there in the choices part? \\ \midrule

        Perception Question (coarse-grained) & \textit{[IMG]} You are given a puzzle. The puzzle consists of a question part on the top and the choices part on the bottom. The question part on the top contains some grids, with the last missing blank grid to be completed. The choices part on the bottom contains a sequence of grids representing the possible choices. How many grids, including the blank grid, are there in the whole puzzle? \\

        \bottomrule
    \end{tabularx}
\caption{
Prompt examples for our experiments.
}
\label{tab:prompt_cot}
\end{table*}

\end{document}